\pdfoutput=1   
\RequirePackage[svgnames,table]{xcolor}
\PassOptionsToPackage{hypertexnames=false}{hyperref}

\documentclass[11pt,letterpaper]{yalearxiv}

\usepackage{graphicx}
\usepackage{float,epstopdf}
\usepackage{bbm}

\usepackage{microtype}

\usepackage[numbers,sort&compress]{natbib}
\setcitestyle{square}

\usepackage{subcaption}
\usepackage{booktabs}

\usepackage{amsmath}
\usepackage{amssymb}
\usepackage{mathtools}
\usepackage{amsthm}
\usepackage{dsfont}
\usepackage{multicol}
\usepackage{makecell}
\usepackage{multirow} 
\usepackage{amsfonts} 
\usepackage{mathrsfs}
\usepackage[amssymb, thickqspace]{SIunits}
\usepackage{enumitem}
\usepackage{pgfplotstable}
\usepackage{lipsum}		

\usepackage{microtype}
\usepackage{graphicx}
\usepackage{booktabs} 
\usepackage[table]{xcolor}
\usepackage{arydshln}
\usepackage[normalem]{ulem} 

\usepackage{cases}
\usepackage{wrapfig}

\usepackage{url}

\usepackage{thmtools}
\usepackage{thm-restate}
\usepackage{tabu}

\definecolor{huskypurple}{HTML}{4B2E83}

\usepackage{titletoc}

\usepackage{listings}
\lstdefinestyle{promptstyle}{
  basicstyle=\ttfamily\footnotesize,
  breaklines=true,
  breakautoindent=false,
  breakindent=0pt,
  postbreak=\mbox{\textcolor{gray}{$\hookrightarrow$}\space},
  columns=fullflexible,
  keepspaces=true,
  frame=single,
  framesep=5pt,
  xleftmargin=6pt,
  xrightmargin=6pt,
  aboveskip=8pt,
  belowskip=8pt,
  showstringspaces=false,
}

\usepackage[capitalize]{cleveref}
\crefname{section}{Sec.}{Secs.}
\Crefname{section}{Section}{Sections}
\Crefname{table}{Table}{Tables}
\crefname{table}{Tab.}{Tabs.}

\newcommand{\cmark}{\ensuremath{\checkmark}}
\newcommand{\xmark}{\ensuremath{\times}}   
\makeatletter
\def\munderbar#1{\underline{\sbox\tw@{$#1$}\dp\tw@\z@\box\tw@}}
\makeatother

\AddToHook{cmd/appendix/before}{%
  \setcounter{axiom}{0}%
}

\newcommand{\be}{\begin{equation}}
\newcommand{\ee}{\end{equation}}
\newcommand{\bea}{\begin{equation*}\begin{aligned}}
\newcommand{\eea}{\end{aligned}\end{equation*}}

\usepackage{fontawesome5}   

\title{CHOREO: From Single Skill to Continuous Humanoid Execution}
\runningtitle{CHOREO: From Single Skill to Continuous Humanoid Execution}

\author{Ziyi~Sun\textsuperscript{1,4,*},
Jingwen~Chen\textsuperscript{1,*},
Yuxi~Wang\textsuperscript{1,\dag,\ddag},
Xiuze~Xia\textsuperscript{1,\dag},\newline
Long~Cheng\textsuperscript{2,3},
Zhaoxiang~Zhang\textsuperscript{2,3},
Junyu~Dong\textsuperscript{1}\\
\faGlobe~\textbf{Project page:} \href{https://ziyisun85-ops.github.io/Project-choreo/}{\texttt{https://ziyisun85-ops.github.io/Project-choreo/}}
}

\firstpagenote{%
\textsuperscript{1}Ocean University of China\quad
\textsuperscript{2}University of Chinese Academy of Sciences\\
\textsuperscript{3}Institute of Automation, Chinese Academy of Sciences\quad
\textsuperscript{4}RoboParty Lab\\
\textsuperscript{*}Equal contribution.\quad
\textsuperscript{\dag}Corresponding authors: \texttt{\{yuxi.wang,xiaxiuze\}@ouc.edu.cn}.\quad
\textsuperscript{\ddag}Project leader.}

\hypersetup{colorlinks=true, linkcolor=blue!50!black, citecolor=blue!50!black,
            urlcolor=blue!50!black,
            pdftitle={CHOREO: From Single Skill to Continuous Humanoid Execution},
            pdfauthor={Ziyi Sun, Jingwen Chen, Yuxi Wang, Xiuze Xia, Long Cheng,
                       Zhaoxiang Zhang, Junyu Dong}}

\begin{document}

\begin{abstract}
Motion data, RL policies, and generative models give humanoids thousands of skills, and the best general trackers execute nearly all of them alone. Chained to follow a language instruction, the skills fail far more often: in simulation, even the best of eight trackers reaches only 84.9\% sequence success, and their average is 48.9\%. This is because validating a skill alone certifies it only for the tracker that ran it and the state it started from, whereas a sequence keeps one tracker and starts each skill wherever the last one ended. Prior work meets these two conditions mostly by training; CHOREO restores them without retraining by treating every skill as a trajectory that any tracker can run. It assigns each instruction the most precise of its best-validated frozen trackers, which ends each skill closer to the next one's start. It then builds each transition from the measured robot state, escalating to bridges and standing only when needed. On 400 instructions, CHOREO lifts sequence success to 95.8\% and beats always standing between skills by 2.8 points with 42\% shorter transitions. On a physical Unitree G1, all 45 zero-shot trials of nine five-action instructions succeed.
\end{abstract}

\maketitle

\begin{figure}[!h]
  \centering
  \includegraphics[width=\textwidth]{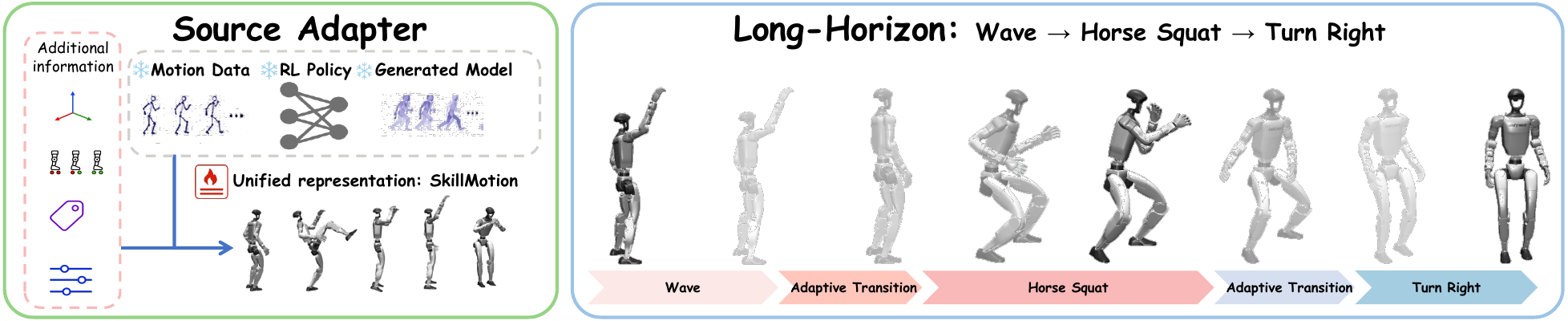}
  \caption{\textbf{CHOREO follows long-horizon instructions by composing skills validated one at a time.} \textbf{Left:} source adapters convert motion data, RL rollouts, and generated motions, with metadata such as coordinate frames, foot contacts, and semantic labels, into SkillMotion assets. \textbf{Right:} for ``wave, hold a horse squat, then turn right,'' the language planner maps the instruction to three actions; CHOREO resolves each to an asset, assigns one frozen tracker to the whole sequence, and builds each transition from the measured robot state.}
  \label{fig:teaser}
\end{figure}

\section{Introduction}
\label{sec:intro}
Humanoids can draw on thousands of reference motions~\cite{mdm,textop,twist2}, and physics-based trackers execute nearly all of them alone~\cite{deepmimic,phc,sonic,holomotion1}. For example, SONIC and SuperMimic succeed in over 99\% of 129,514 benchmark motions~\cite{supermimic}. On our 3,102 skills, two of eight trackers also succeed in over 99\% of single-skill rollouts (\cref{sec:single-asset}). However, skills chained into an instruction such as ``wave, hold a horse squat, then turn right'' (\cref{fig:teaser}) fail far more often: with a single boundary, our two strongest trackers fail 8.4\% and 10.8\% of trials (\cref{tab:assignment}). Which tracker runs the skills matters. Among the trackers with the best validation record on the instruction's skills (\cref{sec:routing}), a random choice fails 12\% of simulated trials and the most precise one fails 4\%. How the skills are joined matters as much: with a Hermite transition at every boundary, the most precise tracker fails 13\% (\cref{tab:assignment,tab:transition-exec}).

Validation is conditional: it certifies a skill only for the tracker that ran it and the state it started from. A sequence runs every skill on one tracker and starts each skill wherever the previous one ended. The robot's velocity, tilt, and foot support at that point depend on how well the previous skill was tracked. Passing validation on every skill is therefore not sufficient, and entry states cannot all be validated in advance, because they depend on every preceding skill.

Prior work meets these conditions mostly by training, if at all (\cref{sec:related}). General trackers train one policy to run every skill~\cite{omnih2o,sonic,gmt,holomotion1}, and physics-based composition retrains the tracker on the composed motion~\cite{closd,stamp,textop}. Skill chaining makes each skill end where the next one succeeds~\cite{burridge1999sequential,konidaris2009skillchaining,lee2019transition,lee2021tstar}, while motion graphs and motion matching connect nominal endpoints~\cite{motiongraphs,holden2020motionmatching}. Language planners choose neither the tracker nor the entry~\cite{ahn2022saycan,liang2023codeaspolicies}.

CHOREO restores both conditions without retraining any tracker or source policy. It stores every skill as a 29-joint trajectory, called a SkillMotion, together with its action, entry and exit boundaries, and validation evidence (\cref{fig:teaser}, left). A reinforcement-learning (RL) policy thus enters as a recorded rollout, and feedback control stays with the tracker. Because every tracker accepts every trajectory, all trackers can be validated on the same skills and compared. CHOREO assigns each instruction the frozen tracker that succeeds most often in validation on all of its skills, and it breaks the frequent ties by tracking precision. A precise tracker leaves the robot closer to the start of each next skill and fails less often at the same entry error (\cref{sec:exp-assignment}). Because a transition can be computed when its boundary is reached, CHOREO builds each transition from the measured robot state and falls back to bridges and standing only when needed (\cref{fig:teaser}, right).
Our contributions can be summarized as follows:
\begin{itemize}
\item We show that skills validated alone fail far more often in sequence because validation is conditional on the tracker and the start state.
\item We propose \textbf{CHOREO}, which restores both conditions without retraining: SkillMotion lets any tracker run any skill, evidence-guided assignment picks each instruction's tracker, and state-aware composition starts each transition from the measured robot state.
\item We evaluate CHOREO in simulation and on a physical Unitree G1. It raises sequence success from 84.9\% (best fixed tracker) to 95.8\%, and breaking validation ties by precision adds 7.6 points. Its composition beats always standing between skills by 2.8 points with 42\% shorter transitions, and all 45 zero-shot real-robot trials succeed.
\end{itemize}

\section{Related Work}
\label{sec:related}
\begin{table}[tbp]
\centering
\caption{Representative systems on the two conditions of single-skill validation (\cref{sec:intro}) and on scope. Skills as references: a separate tracker executes skills as reference motions. Executor chosen by evidence: the tracker or policy is chosen from its execution record. \cmark: shown in the cited paper. \xmark: not found.}
\label{tab:conditions}
\footnotesize
\setlength{\tabcolsep}{3.4pt}
\newcommand{\rh}[2]{{\scriptsize\shortstack{#1\\#2}}}
\begin{tabular}{@{}lccccccc@{}}
\toprule
& \multicolumn{3}{c}{Tracker} & Start & \multicolumn{3}{c}{Scope}\\
\cmidrule(lr){2-4}\cmidrule(lr){5-5}\cmidrule(l){6-8}
System & \rh{Skills as}{references} & \rh{No}{retraining} & \rh{Executor chosen}{by evidence} & \rh{Transition from}{measured state} & \rh{Mixed skill}{sources} & \rh{Object}{interaction} & \rh{Real}{humanoid}\\
\midrule
GMT~\cite{gmt} & \cmark & \cmark & \xmark & \xmark & \cmark & \xmark & \cmark\\
Switch~\cite{switch} & \cmark & \xmark & \xmark & \cmark & \xmark & \xmark & \cmark\\
\midrule
TextOp~\cite{textop} & \cmark & \cmark & \xmark & \xmark & \cmark & \xmark & \cmark\\
CLoSD~\cite{closd} & \cmark & \xmark & \xmark & \cmark & \xmark & \cmark & \xmark\\
Transition policies~\cite{lee2019transition} & \xmark & \xmark & \xmark & \cmark & \xmark & \cmark & \xmark\\
\midrule
SayCan~\cite{ahn2022saycan} & \xmark & \cmark & \xmark & \xmark & \cmark & \cmark & \xmark\\
RAX~\cite{rax} & \cmark & \cmark & \xmark & \cmark & \cmark & \cmark & \xmark\\
RoboHarness~\cite{roboharness} & \xmark & \cmark & \cmark & \cmark & \cmark & \cmark & \xmark\\
\midrule
\textbf{CHOREO} & \cmark & \cmark & \cmark & \cmark & \cmark & \xmark & \cmark\\
\bottomrule
\end{tabular}
\end{table}

We compare prior work on the two conditions under which single-skill validation holds (\cref{sec:intro}): the tracker that runs a skill and the state it starts from (\cref{tab:conditions}).

\subsection{Motion Tracking and Whole-Body Control}
\label{sec:rw-tracking}
Physics-based imitation~\cite{deepmimic,phc} has scaled to general trackers that follow large motion sets with one policy~\cite{gmt,sonic,supermimic,bfmzero,twist2,any2track,fromw1,holomotion1,protomotions,humanoidgpt,mimiclite}. Skill-based controllers run a set of skills and their transitions through one policy~\cite{langwbc,spectralskills,metatron,switch,omnicontact,humanoidhanoi}. Both train one controller to run every skill, so its training limits what it can run. Untrackable or infeasible motions are removed from the training data~\cite{gmt,textop,supermimic}, and untrained skills cannot run~\cite{switch}. On our benchmark, the eight trackers reach 6.35--84.85\% sequence success, and the choice matters even among trackers that pass validation on every skill of an instruction (\cref{tab:assignment}). CHOREO therefore keeps all trackers frozen and assigns each instruction the tracker with the best validation record on its skills.

\subsection{Motion Generation and Temporal Composition}
\label{sec:rw-generation}
Text-driven generators synthesize clips and long sequences~\cite{mdm,ardy,doubletake,textop,omg}. Clip-based methods join existing motions by graphs, matching, or in-betweening~\cite{motiongraphs,holden2020motionmatching,php,motionbricks,stamp}. Both plan each transition from reference poses, which assumes that the robot starts each skill on its reference. In practice, a robot starts each skill wherever tracking error left it. In our ablation, planning from the reference instead of the measured state lowers sequence success at every instruction length (\cref{sec:exp-ablation}). Other methods handle this gap in advance: they train the tracker on the composed motion~\cite{php,stamp,closd,textop}, or design skills that end where the next one succeeds~\cite{burridge1999sequential,konidaris2009skillchaining,lee2019transition,lee2021tstar}. CHOREO needs no training. It builds each transition from the measured robot state.

\subsection{Language Planning and Skill Orchestration}
\label{sec:rw-planning}
Language planners use LLMs to break an instruction into skills and choose the next one~\cite{ahn2022saycan,innermonologue,liang2023codeaspolicies}. Agent frameworks add whole-body commands, code skills, or routing among policies by execution evidence~\cite{hitch,rax,roboharness}. These systems decide which skill runs next, and some also shape how the robot enters it within one task program~\cite{rax} or command~\cite{hitch}. The closest, RoboHarness~\cite{roboharness}, picks arm policies by execution evidence and bridges each handoff from the robot's state. CHOREO does both for humanoid skills: it picks a frozen tracker and builds each transition.

\section{Method}
\label{sec:method}
\paragraph{Problem setting.}
An instruction lists actions in order. CHOREO executes each action with an asset, which is a skill stored as a SkillMotion trajectory (\cref{sec:representation}). Given the assets $a_{1:m}$ and a set $\mathcal K$ of frozen trackers, CHOREO makes two decisions: one tracker $k^\star\in\mathcal K$ for the whole instruction, and a transition at each of the $m-1$ boundaries between consecutive assets. A trial succeeds if the robot completes all $m$ assets in order (\cref{sec:setup}).

\begin{figure}[!ht]
  \centering
  \includegraphics[width=\textwidth]{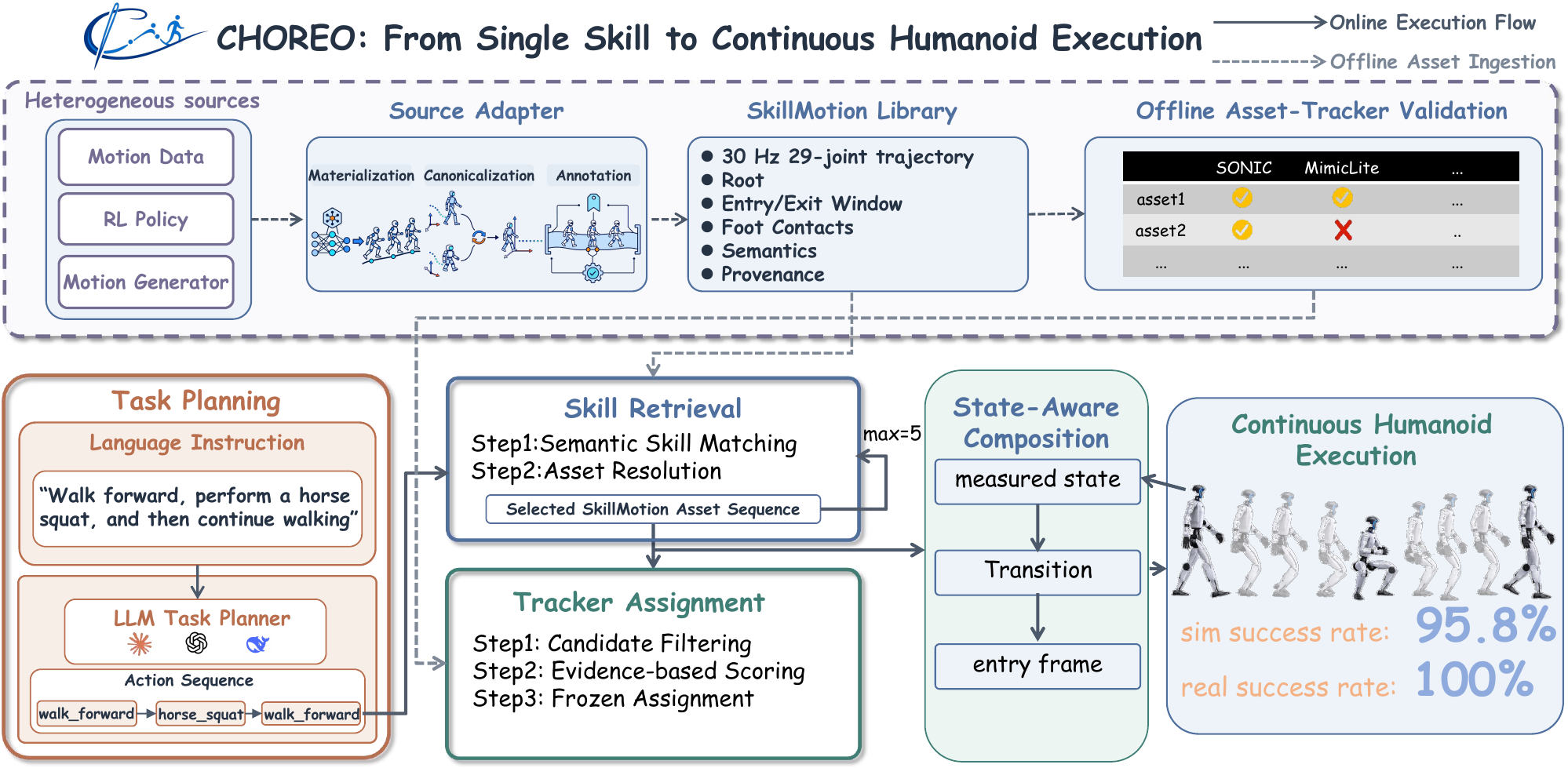}
  \caption{Overview of CHOREO. Offline, source adapters convert motion data, RL rollouts, and generated motions into SkillMotion assets, and every candidate tracker is validated on 1,010 of them. Online, the language planner maps the instruction to actions and the runtime resolves each action to an asset. Evidence-guided assignment then selects one frozen tracker, and state-aware composition builds each transition from the measured robot state.}
  \label{fig:overview}
\end{figure}

\paragraph{Overview.}
Single-skill validation runs each asset alone, so its result holds only for the tracker that ran the asset and the state the asset started from (\cref{sec:intro}). In a sequence, however, one tracker runs all assets, and each asset starts where the previous one ended. CHOREO handles these two conditions with two modules (\cref{fig:overview}). Tracker assignment (\cref{sec:routing}) chooses a tracker that passed validation on every asset of the instruction, taking the most precise one among those with the highest validation success. A precise tracker also leaves the robot closer to the start of each next asset. State-aware composition (\cref{sec:composition}) builds each transition from the measured robot state and enters the next asset near its start, as in validation.

\subsection{SkillMotion Library and Language Planning}
\label{sec:representation}
\paragraph{SkillMotion.}
CHOREO stores every skill as a reference trajectory, so any tracker can run it and transitions can be computed online. Source adapters convert motion data, recorded RL rollouts, and offline outputs of generative models into such SkillMotion assets. They retarget motions to the G1 when necessary and check only the data format, not executability. Each asset $a$ stores the trajectory and the information that the two decisions need:
\begin{equation}
\begin{split}
\mathcal M_a&=\big(\{\mathbf Q_t,\mathbf V_t\}_{t=1}^{N},\,A(a),\,\mathcal B_a,\,\mathcal E_a\big),\\
\mathbf Q_t&=[p_t,r_t,\theta_t],\qquad
\mathbf V_t=[v_t,\omega_t,\dot\theta_t].
\end{split}
\label{eq:skillmotion}
\end{equation}
The trajectory has $N$ frames at $f=30$~Hz. Here $p_t\in\mathbb R^3$ is the world root position, $r_t\in S^3$ the root orientation, $\theta_t\in\mathbb R^{29}$ the named G1 joint angles, and $v_t$, $\omega_t$, $\dot\theta_t$ their velocities. Besides the action $A(a)$, the asset stores two records. The validation evidence $\mathcal E_a$ records how each candidate tracker executes $a$ alone (\cref{sec:routing}). The boundary descriptor $\mathcal B_a$ marks the low-speed entry and exit windows, where $a$ can be entered and left, and stores estimated foot contacts with confidence (\cref{sec:composition}).

\paragraph{Language planning.}
A frozen external LLM maps the instruction to action identifiers from a catalog of validated actions (\cref{sec:appendix-language}). The runtime then resolves each action to a validated asset, either an explicitly specified one or the action's default.

\subsection{Evidence-Guided Tracker Assignment}
\label{sec:routing}
\paragraph{Validation evidence.}
Offline, 1,010 of the 3,102 library assets are each run alone with every candidate tracker in three validation rollouts, judged by the trial success criterion (\cref{sec:setup}). Each candidate tracker is a frozen policy with its native observations, control rate, and actuator mapping. For asset $a$ and tracker $k$, the evidence $\mathcal E_a$ stores the number $s_{ak}$ of successful rollouts out of $n_{ak}=3$ and the validation MPJPE $\epsilon_{ak}$, averaged over all three rollouts, including failed ones. Tracker $k$ is eligible for $a$ ($k\in K_a$) if $s_{ak}\geq1$.

\paragraph{Assignment rule.}
One tracker runs every asset, and a trial fails if any asset fails. CHOREO therefore keeps only the trackers that are eligible for all assets. It ranks them by validation success. However, success alone is not enough, so CHOREO breaks ties by precision (lower validation MPJPE):
\begin{equation}
K(a_{1:m})={\textstyle\bigcap_{i=1}^{m}}K_{a_i},\qquad
k^\star=\operatorname*{arg\,min}_{k\in K(a_{1:m})}\big(-\hat p_k,\ \bar\epsilon_k\big).
\label{eq:assignment}
\end{equation}
Here $\hat p_k=\sum_i s_{a_ik}/\sum_i n_{a_ik}$ is the validation success pooled over the $m$ assets, $\bar\epsilon_k$ is the mean of $\epsilon_{a_ik}$, and the pair is compared lexicographically. On 302 of the 400 benchmark instructions, several trackers tie in validation success, so precision often decides. Precision helps at the boundaries, because a precise tracker reaches each next asset with a smaller entry error (the root-relative position error at its entry frame) and also fails less often at the same entry error (\cref{sec:exp-assignment}). If no tracker is eligible for all assets, the trial counts as failed. Different instructions can use different trackers, and no tracker or source policy is updated.

\begin{table*}[tbp]
\centering
\caption{Evidence-guided tracker assignment versus eight fixed trackers and two selection baselines. The table gives SSR and sequence velocity error by instruction length $L$, and MPJPE. Random eligible picks uniformly among the trackers eligible for every asset of the instruction. Validation success ranks these trackers by pooled validation success alone and breaks ties uniformly, whereas CHOREO breaks them by validation MPJPE (\cref{eq:assignment}). Both baseline rows average over their random choices. Each length has 100 instructions with five paired seeds (500 trials), and All has 2,000. The CHOREO row equals that of \cref{tab:transition-exec}. $\uparrow$/$\downarrow$: higher/lower is better. Bold/underline: best/second-best per column.}
\label{tab:assignment}
\footnotesize
\setlength{\tabcolsep}{4pt}
\begin{tabular}{lrrrrrcccccc}
\toprule
& \multicolumn{5}{c}{SSR (\%) $\uparrow$} & \multicolumn{5}{c}{Seq. velocity error (m/s) $\downarrow$} & \\
\cmidrule(lr){2-6}\cmidrule(lr){7-11}
Tracker / assignment & $L=2$ & $3$ & $5$ & $8$ & All & $L=2$ & $3$ & $5$ & $8$ & All & MPJPE (m) $\downarrow$\\
\midrule
SONIC & 91.60 & 90.60 & 82.20 & 69.40 & 83.45 & \underline{0.145} & \textbf{0.111} & \underline{0.123} & \textbf{0.129} & \underline{0.127} & \textbf{0.0278}\\
MimicLite & 77.00 & 67.20 & 44.40 & 24.20 & 53.20 & 0.207 & 0.158 & 0.169 & 0.168 & 0.176 & 0.0459\\
TWIST2 & 44.00 & 29.00 & 5.40 & 1.00 & 19.85 & 0.423 & 0.387 & 0.373 & 0.387 & 0.393 & 0.1265\\
OpenTrack & 8.60 & 12.60 & 4.20 & 0.00 & 6.35 & 0.504 & 0.433 & 0.424 & 0.429 & 0.448 & 0.2351\\
FRoM-W1 & 65.60 & 68.40 & 46.40 & 28.60 & 52.25 & 0.366 & 0.338 & 0.321 & 0.347 & 0.343 & 0.1552\\
HoloMotion & 89.20 & 87.00 & 82.00 & \underline{81.20} & 84.85 & 0.194 & 0.187 & 0.171 & 0.168 & 0.180 & 0.0489\\
ProtoMotions GTP & 59.40 & 50.00 & 28.40 & 10.00 & 36.95 & 0.276 & 0.238 & 0.239 & 0.250 & 0.251 & 0.0574\\
Humanoid-GPT & 70.20 & 74.80 & 49.60 & 23.80 & 54.60 & 0.191 & 0.149 & 0.142 & 0.146 & 0.157 & 0.0326\\
\midrule
Random eligible & 85.69 & 83.21 & 72.67 & 71.46 & 78.26 & 0.214 & 0.189 & 0.185 & 0.193 & 0.195 & 0.0619\\
Validation success & \underline{93.60} & \underline{92.49} & \underline{85.49} & 81.07 & \underline{88.17} & 0.184 & 0.154 & 0.157 & 0.167 & 0.166 & 0.0496\\
\midrule
CHOREO & \textbf{100.00} & \textbf{97.80} & \textbf{96.40} & \textbf{89.00} & \textbf{95.80} & \textbf{0.130} & \underline{0.113} & \textbf{0.115} & \underline{0.136} & \textbf{0.123} & \underline{0.0280}\\
\bottomrule
\end{tabular}
\end{table*}

\subsection{State-Aware Motion Composition}
\label{sec:composition}
At each boundary, the robot has finished an asset $a$ in the measured state $x$ and must enter the next asset $a'$. The measured state differs from the nominal end of $a$ by the tracking error so far, so CHOREO starts the transition at $x$. Because $a'$ was validated from its start, CHOREO enters it at the entry-window frame closest to $x$, and it uses longer transitions only when no frame is close enough.

\paragraph{Entry matching.}
For each frame $j$ in the entry window $\mathcal W^{\rm in}_{a'}$, the boundary error $\mathbf e(x,b'_j)\in\mathbb R^8$ compares $x$ with the reference state $b'_j$ of $a'$ at that frame. It covers posture (joint angles and velocities), root motion (height, tilt, horizontal offset, and linear and angular velocity), and support (foot contact). Entry matching selects the frame with the lowest continuity score $\mathbf w^\top\mathbf e$, and the boundary is compatible if every error term of this frame is within its threshold:
\begin{equation}
\begin{split}
j^\star&=\operatorname*{arg\,min}_{j\in\mathcal W^{\rm in}_{a'}}\mathbf w^\top\mathbf e(x,b'_j),\\
\text{compatible}&\iff\mathbf e(x,b'_{j^\star})\leq\boldsymbol\tau\ \text{componentwise}.
\end{split}
\label{eq:boundary-search}
\end{equation}
The weights $\mathbf w$ and thresholds $\boldsymbol\tau$ are fixed defaults, set without held-out tuning (\cref{tab:boundary-params}).

\paragraph{Transitions from the measured state.}
Every transition starts at $x$, and the asset it enters is re-anchored so that its entry frame lies at the robot's measured position and heading. There are three kinds of transitions. A Hermite transition interpolates from $x$ to the entry frame over 24 frames (0.8~s) with a cubic Hermite spline that matches the measured velocities at the start and the reference velocities at the end. It is the shortest transition, but it suits only states close to the entry frame. A bridge chain inserts existing assets from a pool of eight bridge actions. Bridge search adds one bridge at a time, each time picking the bridge with the lowest entry and exit scores plus a duration penalty among those it can enter compatibly. It stops once the chain can enter $a'$ compatibly and fails if more than five bridges are needed. The standing fallback (\textsf{Stand}) moves the robot into the last (at most 120) frames of the standing bridge and then into $a'$, using 24-frame quintic transitions. It takes several seconds but returns the robot to standing, the state in which every trial starts.

\paragraph{Escalation.}
The fixed rules of \cref{sec:exp-transition} treat every boundary the same way. CHOREO instead uses the shortest transition it accepts at each boundary:
\begin{equation}
D=\begin{cases}
\textsf{Hermite}, & \text{compatible}\ \lor\ c^\star<\bar c,\\
\textsf{Bridge}, & \text{otherwise, if bridge search succeeds},\\
\textsf{Stand}, & \text{otherwise}.
\end{cases}
\label{eq:adaptive}
\end{equation}
Here $c^\star$ is the continuity score of the matched entry frame. The condition $c^\star<\bar c$, called the adaptive decision, accepts a Hermite transition even when some error term exceeds its threshold, as long as $c^\star$ stays below the cutoff $\bar c$.

\section{Experiments}
\label{sec:experiments}
Each experiment answers one question. The first three test the two conditions of \cref{sec:intro}, and all but the last run in simulation. \textbf{RQ1:} Among trackers eligible for every asset of an instruction, does the choice matter, and does evidence-guided assignment beat every fixed tracker? \textbf{RQ2:} With the tracker fixed, does state-aware composition beat fixed transition rules? \textbf{RQ3:} Which components of the composition matter, including starting from the measured state? \textbf{RQ4:} How robust is the composition to velocity impulses before boundaries? \textbf{RQ5:} Does CHOREO transfer zero-shot to a physical G1 (\cref{fig:hardware})?

\subsection{Experimental Setup}
\label{sec:setup}
\begin{wraptable}[14]{r}{0.51\textwidth}
\centering
\vspace{-8pt}
\caption{Reference transitions before tracking. The columns give the mean joint-angle jump $J_\theta$ (rad) and joint-velocity jump $J_{\dot\theta}$ (rad/s) at a boundary, the mean transition duration (s), and the transition share of successful-trial time (\%). Bold/underline: lowest/second-lowest jumps.}
\label{tab:transition-ref}
\footnotesize
\begin{tabular}{lrrrr}
\toprule
& \multicolumn{2}{c}{Reference jump $\downarrow$} & \multicolumn{2}{c}{Transition}\\
\cmidrule(lr){2-3}\cmidrule(lr){4-5}
Method & $J_\theta$ & $J_{\dot\theta}$ & Time & Share\\
\midrule
Direct Concatenation & 0.3358 & 0.438 & 0.00 & 0.0\\
Fixed Hermite & 0.0421 & 0.314 & 0.78 & 8.7\\
Fixed Stand Bridge & \textbf{0.0325} & \textbf{0.212} & 5.54 & 39.0\\
Motion Matching & 0.0378 & 0.237 & 5.20 & 40.2\\
CHOREO & \underline{0.0342} & \underline{0.216} & 3.24 & 29.0\\
\bottomrule
\end{tabular}
\vspace{-6pt}
\end{wraptable}
\paragraph{Assets and instructions.}
The library contains 3,102 G1 assets: 2,950 from motion data, 76 recorded RL rollouts, and 76 generated motions. The benchmark has 100 instructions at each length $L\in\{2,3,5,8\}$, covering 100 distinct actions in 1,800 action slots. Five paired seeds per instruction give 500 trials per length and 2,000 per method. All methods share instructions, prompts, resolved assets, initialization, and metrics, and they differ only in the tracker or the composition rule. The language planner recovers the ground-truth action sequence for all 400 instructions, so planning errors affect no comparison.

\paragraph{Trackers.}
The eight candidates are SONIC~\cite{sonic}, MimicLite~\cite{mimiclite}, TWIST2~\cite{twist2}, OpenTrack (open-source implementation of~\cite{any2track}), FRoM-W1~\cite{fromw1} with its H-ACT controller, HoloMotion~\cite{holomotion1}, ProtoMotions GTP~\cite{protomotions}, and Humanoid-GPT~\cite{humanoidgpt}. Each keeps its native model, observations, PD gains, control rate, and termination settings (\cref{tab:native}), so the comparisons measure complete systems rather than retrained variants. The validation evidence (\cref{sec:routing}) comes only from running each asset alone and is never updated with test outcomes.

\paragraph{Metrics.}
A trial succeeds if every action completes in order without a fall, timeout, or abnormal termination, and if $E_{\max}=\max_t e_t\leq0.3$~m as in TextOp~\cite{textop}. Here $e_t$ is the mean root-relative position error over 23 non-root body links. Sequence success rate (SSR) and fall rate (FR) divide by all trials, and transition success rate (TSR) divides by attempted boundaries. MPJPE averages $e_t$ over the observed frames of each trial, including failed trials, and $E_v$ is its velocity analogue. $J_\theta$ and $J_{\dot\theta}$ are the mean joint-angle and joint-velocity jumps across a boundary, in the reference (\cref{tab:transition-ref}) and in the executed motion (\cref{tab:transition-exec}; \cref{sec:metrics}).

\paragraph{Statistics.}
All comparisons are paired by instruction and seed. Experiments 1 and 2 compute confidence intervals and $p$ values by resampling whole instructions, and Experiments 3 and 4 use McNemar tests. All $p$ values are Holm-corrected within each family of comparisons against CHOREO.

\subsection{Experiment 1: Tracker Assignment}
\label{sec:exp-assignment}
To answer RQ1, we compare the assignment rule of \cref{eq:assignment} with each of the eight fixed trackers and with two rules that ignore precision. Random eligible picks uniformly among the eligible trackers, and Validation success ranks them by validation success alone. All methods use CHOREO's composition and the same assets and instruction/seed pairs. Since the validation evidence covers every asset that the instructions use, this experiment tests how well single-skill evidence predicts sequence execution, not generalization to unvalidated assets. Random eligible and Validation success average over their random choices.

\paragraph{Results.}
The choice among eligible trackers matters at both levels of the rule (\cref{tab:assignment}). Random eligible reaches 78.26\% SSR, below the best fixed tracker. Ranking by validation success raises SSR to 88.17\%, and breaking its ties by precision raises it to 95.80\%. Validation success saturates: it decides the assignment of only 79 instructions, while on 302 several trackers share the highest success and validation MPJPE breaks the tie. On all other instructions, success-only ranking picks the same tracker as CHOREO, so CHOREO's gain of 7.63 percentage points (pp) over it comes entirely from choosing the most precise of the tied trackers. This gain cuts failures from 11.83\% to 4.20\%. The gain over random choice is 17.54~pp, and both gains have 95\% confidence intervals that exclude zero (Holm-adjusted $p=0.0003$). Precision helps at the boundaries (\cref{sec:assignment-analyses}). On the 302 tied instructions, the most precise tracker enters each next asset with smaller errors and fails less often after each boundary. It succeeds in 98.15\% of trials, against 88.08\% for the second most precise, and the gap grows from 5.06~pp at $L=2$ to 18.93~pp at $L=8$. Over these instructions, validation MPJPE correlates negatively with sequence success (Spearman $\rho=-0.32$).

No fixed tracker suffices either. SONIC leads for $L\leq5$ but drops to 69.40\% at $L=8$, where HoloMotion is strongest (81.20\%), and the other six stay below 55\% overall. Evidence-guided assignment reaches 95.80\%, a gain of 10.95~pp (95\% CI [7.55, 14.30]) over the best fixed tracker, HoloMotion (84.85\%). At each length, it beats the strongest fixed tracker by 8.40, 7.20, 14.20, and 7.80~pp ($L=2,3,5,8$), and it succeeds in all 500 trials at $L=2$. Only one instruction ($L=3$) has an empty eligible set, and its five trials count as failures. CHOREO's MPJPE (0.0280~m) is within 0.2~mm of the most precise fixed tracker, SONIC, so the gain does not cost tracking precision. Success still declines with length, to 89.00\% at $L=8$, where every failed trial is a fall (\cref{sec:exp-transition}). Averaged over tracker subsets of each size, the SSR of random choice among eligible trackers levels off near 78\%, whereas CHOREO's SSR keeps rising, from 46.84\% with one tracker to 95.80\% with all eight (\cref{tab:pool-size}).

\subsection{Experiment 2: Motion Composition}
\label{sec:exp-transition}
\begin{figure*}[tp]
  \centering
  \includegraphics[width=\textwidth]{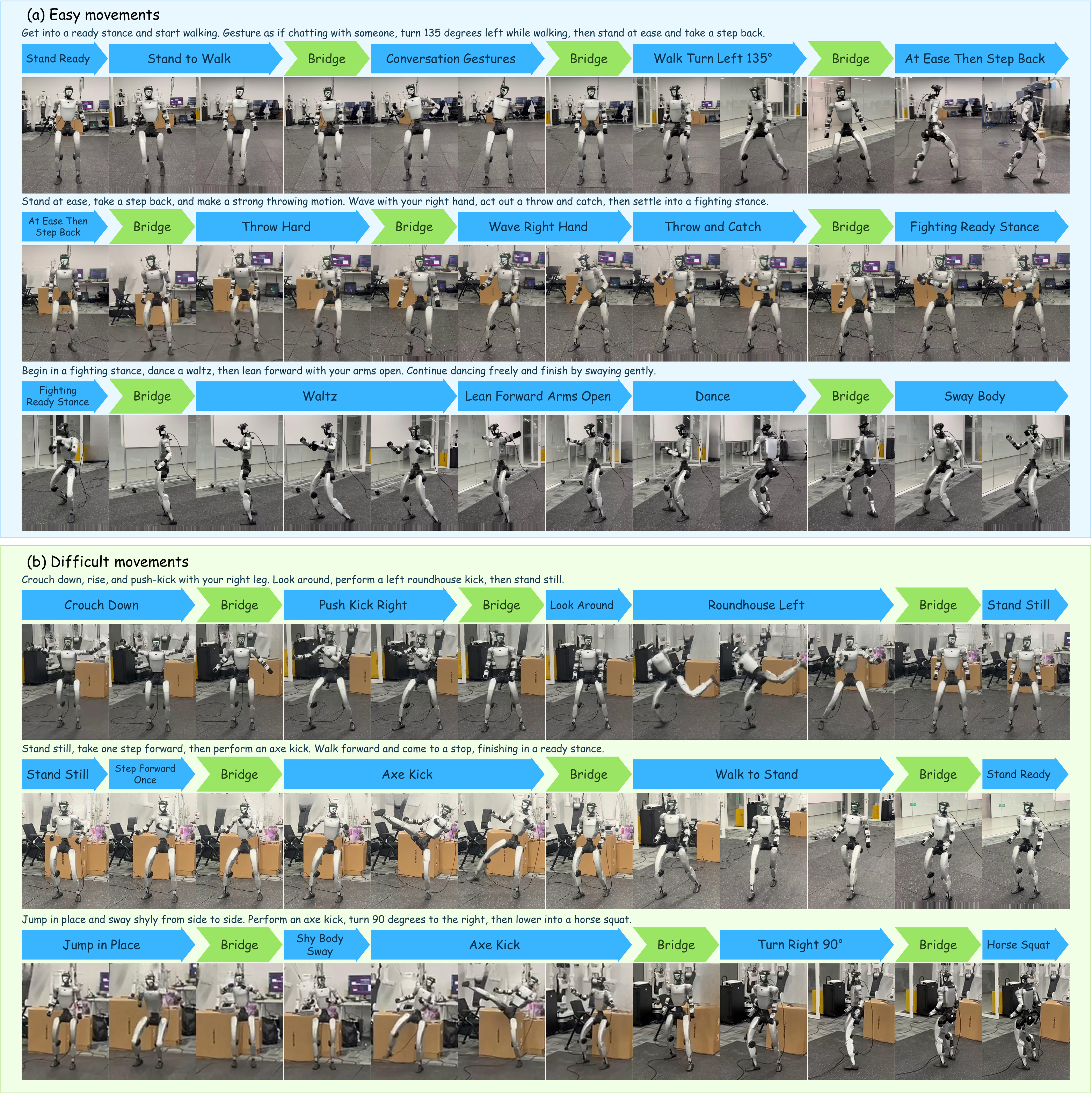}
  \caption{Zero-shot deployment of CHOREO on a physical Unitree G1. (a) Three of the six easy instructions. (b) The three hard instructions. All were chosen among instructions that succeed in simulation. Each row shows the five actions (blue), the bridges inserted at boundaries (green), and video frames. Boundaries without a bridge use a Hermite transition. All 45 trials of the nine instructions succeed.}
  \label{fig:hardware}
\end{figure*}

To answer RQ2, we compare CHOREO's composition (\cref{sec:composition}) with four fixed rules. Direct Concatenation joins consecutive assets without blending. Fixed Hermite always inserts a Hermite transition. Fixed Stand Bridge always routes through the standing bridge. Motion Matching selects boundary frames by phase. All methods share the tracker assignment, assets, and paired seeds, so only the composition rule changes. Each method has 2,000 trials and 7,000 semantic boundaries.

\paragraph{Results.}
No fixed rule suits every boundary (\cref{tab:transition-ref,tab:transition-exec}). Direct Concatenation leaves reference joint-angle jumps 8 to 10 times larger than the other rules (0.336~rad on average), and 15.30\% of its trials fall. Fixed Hermite cuts the executed velocity jump from 2.683 to 1.106~rad/s, but 12.55\% of its trials still fall. Fixed Stand Bridge stands at every boundary and is the strongest baseline. Its reference is the most continuous, and its MPJPE matches CHOREO's (0.0279 vs.\ 0.0280~m). However, its transitions last 5.54~s on average and take 39.0\% of successful-trial time. CHOREO stands only when it accepts neither a Hermite transition nor a bridge chain (\cref{eq:adaptive}). Its reference is nearly as continuous ($J_{\dot\theta}$ of 0.216 vs.\ 0.212~rad/s), and its transitions are 42\% shorter (3.24~s, 29.0\% of the time). It has the highest SSR at every length and overall (95.80\%) and the lowest FR (3.95\% vs.\ 6.20\%). In paired SSR, it beats Direct Concatenation by 12.75~pp, Fixed Hermite by 9.10~pp, Fixed Stand Bridge by 2.75~pp, and Motion Matching by 8.70~pp. All four gains have 95\% confidence intervals that exclude zero (Holm-adjusted $p\leq0.0029$). Its executed velocity jump matches the best baseline (0.938 vs.\ 0.941~rad/s), and it has the lowest peak velocity and accelerations near boundaries.

\begin{table*}[tbp]
\centering
\caption{Execution of the transitions in \cref{tab:transition-ref} in simulation. Every method uses the same per-instruction tracker (500 trials per length, 2,000 per method). TSR divides by attempted boundaries. $J_\theta$ (rad) and $J_{\dot\theta}$ (rad/s) are executed jumps at a boundary. $\uparrow$/$\downarrow$: higher/lower is better. Bold/underline: best/second-best per column.}
\label{tab:transition-exec}
\footnotesize
\setlength{\tabcolsep}{3.3pt}
\begin{tabular}{lrrrrrcccccc}
\toprule
& \multicolumn{5}{c}{SSR (\%) $\uparrow$} & & & \multicolumn{2}{c}{MPJPE (m) $\downarrow$} & \multicolumn{2}{c}{Executed jump $\downarrow$}\\
\cmidrule(lr){2-6}\cmidrule(lr){9-10}\cmidrule(lr){11-12}
Method & $L=2$ & $3$ & $5$ & $8$ & All & TSR (\%) $\uparrow$ & FR (\%) $\downarrow$ & Seq. & Trans. & $J_\theta$ & $J_{\dot\theta}$\\
\midrule
Direct Concatenation & 88.40 & 87.20 & 83.60 & 73.00 & 83.05 & \textbf{99.48} & 15.30 & 0.0406 & 0.0536 & 0.0430 & 2.683\\
Fixed Hermite & 90.60 & 89.40 & 88.80 & 78.00 & 86.70 & 98.86 & 12.55 & 0.0346 & 0.0322 & 0.0183 & 1.106\\
Fixed Stand Bridge & \underline{94.60} & \underline{97.00} & \underline{96.00} & \underline{84.60} & \underline{93.05} & 98.95 & \phantom{0}\underline{6.20} & \textbf{0.0279} & \textbf{0.0245} & \underline{0.0164} & 0.969\\
Motion Matching & 92.20 & 86.40 & 88.60 & 81.20 & 87.10 & 97.00 & 12.45 & 0.0313 & 0.0315 & \textbf{0.0160} & \underline{0.941}\\
CHOREO & \textbf{100.00} & \textbf{97.80} & \textbf{96.40} & \textbf{89.00} & \textbf{95.80} & \underline{99.07} & \phantom{0}\textbf{3.95} & \underline{0.0280} & \underline{0.0251} & \textbf{0.0160} & \textbf{0.938}\\
\bottomrule
\end{tabular}
\end{table*}

\begin{figure*}[t]
\centering
\begin{minipage}[t]{0.485\textwidth}
  \includegraphics[width=\linewidth]{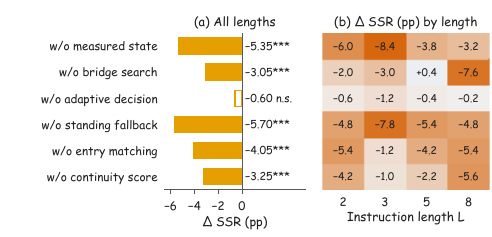}
  \caption{Ablation of the composition rule. Each variant removes one component. (a) Change in SSR from CHOREO (95.80\%) over 2,000 trials. The hollow bar is not significant. (b) Change at each length (500 trials). Tests are McNemar tests on paired trials, Holm-corrected over the six variants. $^{***}p<0.001$.}
  \label{fig:ablation}
\end{minipage}\hfill
\begin{minipage}[t]{0.485\textwidth}
  \includegraphics[width=\linewidth]{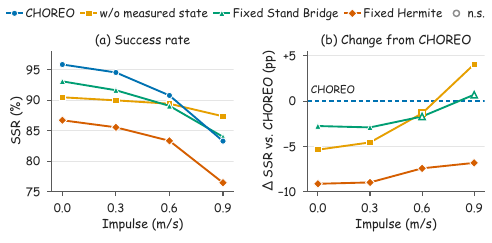}
  \caption{Robustness to a horizontal root velocity impulse before boundaries. Every method uses CHOREO's tracker assignment. (a) SSR over 2,000 trials per point. (b) SSR minus CHOREO's SSR, in pp. Hollow markers are not significant ($p\geq0.05$). Tests are McNemar tests on paired trials, Holm-corrected over the 12 comparisons.}
  \label{fig:robustness}
\end{minipage}
\end{figure*}

TSR is at least 97\% for every method, because it counts only the boundaries that a trial reaches. Apart from the five trials of the unassigned instruction, every failed CHOREO trial is a fall. Of the 79 falls, 55 occur at $L=8$, and 72 follow a standing fallback, which CHOREO uses only at the least compatible boundaries (\cref{sec:transition-extra}). All 766 boundaries between motion data and an RL rollout or a generated motion succeed, covering 149 distinct action pairs, and so do all 544 executed segments of the seven actions that use such assets. Boundaries between two motion-data assets succeed in 98.9\% of 5,972 attempts and account for every failed boundary (\cref{sec:cross-source}).

\subsection{Experiment 3: Component Ablation}
\label{sec:exp-ablation}
To answer RQ3, we remove one component of the composition in each variant and keep everything else fixed (\cref{sec:ablation-appendix}). Every component contributes, the adaptive decision least (\cref{fig:ablation}). Planning transitions from the nominal end of the previous asset instead of the measured state lowers SSR from 95.80\% to 90.45\%, and removing the standing fallback lowers it to 90.10\%, both at every length. Removing entry matching costs 4.05~pp, removing the continuity score 3.25~pp, and removing bridge search 3.05~pp, mainly at $L=8$ ($-7.60$~pp). Removing the adaptive decision costs 0.60~pp, with a small drop at every length.

\subsection{Experiment 4: Robustness to Perturbations}
\label{sec:exp-robustness}
To answer RQ4, we apply a horizontal velocity impulse of 0.3, 0.6, or 0.9~m/s to the robot's root before boundaries. The impulses lower CHOREO's SSR from 95.80\% to 94.50\%, 90.75\%, and 83.30\% (\cref{fig:robustness}). Under small impulses, up to 0.3~m/s, CHOREO stays significantly above all three alternatives: at 0.3~m/s, it keeps 94.50\% SSR, against 91.60\% for Fixed Stand Bridge, 89.95\% without the measured state, and 85.55\% for Fixed Hermite. It also remains above Fixed Hermite at every magnitude. Under larger impulses, Fixed Stand Bridge and the variant without the measured state catch up, and the variant leads at 0.9~m/s (\cref{fig:robustness}b). Handling such large disturbances needs further study.

\subsection{Experiment 5: Real-Robot Deployment}
\label{sec:exp-hardware}
To answer RQ5, we deploy CHOREO zero-shot on a physical Unitree G1, without retraining or retuning. Each instruction runs with the tracker of \cref{eq:assignment}, and every transition starts from the measured robot state. From the $L=5$ instructions that succeed in simulation, we select six easy and three hard ones (\cref{tab:hardware}). A hard instruction contains several dynamic actions or a large whole-body posture change. Each instruction runs five times, which gives 45 trials. A trial succeeds if the robot completes all five actions in order without a fall. All 45 trials succeed, including the 15 hard ones (\cref{fig:hardware}). Because the instructions come from simulated successes, this experiment tests zero-shot transfer, not hardware success over the full benchmark.

\section{Conclusion}
\label{sec:conclusion}
Validation certifies a skill only for the tracker that ran it and the state it started from, so skills that pass alone can still fail in sequence. CHOREO restores both conditions for frozen trackers by treating every skill as a trajectory: it assigns each instruction the most precise of its best-validated trackers and builds each transition from the measured robot state. In simulation, assignment raises sequence success from 84.9\% for the best fixed tracker to 95.8\%, and breaking validation ties by precision is worth 7.6 points. The composition beats always standing between skills by 2.8 points with 42\% shorter transitions, and planning from nominal endpoints instead of the measured state costs 5.4 points. On a physical Unitree G1, all 45 zero-shot trials succeed.

\subsection*{Limitations and Future Work}
\phantomsection
\label{sec:limitations}
\begin{itemize}
\item \textbf{No terrain or object interaction.} CHOREO observes only the robot's own state. Its references are therefore neither adapted to terrain nor conditioned on objects, and object-related actions such as lifting a box are reproduced as motions without evaluating manipulation. Adding terrain and object states to transition construction and asset selection would lift this restriction.
\item \textbf{No online motion generation.} CHOREO composes only assets in its offline library, including generated ones, so a new action needs offline conversion and validation. Online motion generators with fast validation are a natural next step.
\item \textbf{Manual library screening.} The library's motions and the bridge pool are screened manually. The validation evidence that CHOREO already records could automate this screening.
\end{itemize}

\section*{Acknowledgments}
This work was supported by the National Natural Science Foundation of China under Grant~62606500, and by the Natural Science Foundation of Shandong Province, China, under Grants ZR2026QC1098 and 2026HWYQ-026.

\bibliography{main}
\bibliographystyle{plainnat}

\clearpage
\appendix
{\titlefont Supplementary Material\par}
\vspace{6pt}
\setcounter{table}{0}
\setcounter{figure}{0}
\setcounter{equation}{0}
\renewcommand{\thetable}{S\arabic{table}}
\renewcommand{\thefigure}{S\arabic{figure}}
\renewcommand{\theequation}{S\arabic{equation}}

\section{Representation and Composition Details}
\label{sec:appendix-representation}
\paragraph{Coordinates and sampling.}
SkillMotion uses meters, radians, seconds, named G1 joints, world-frame root translations, and $xyzw$ reference quaternions. The 29-joint order comprises six joints per leg, three waist joints, and seven joints per arm. Conversion to a tracker's native order is explicit. A 23-joint source is mapped to the 29-joint representation by zero-filling the missing wrist joints; this makes the asset compatible with the interface but provides neither wrist motion nor tracking evidence. Besides the fields in \cref{eq:skillmotion}, each stored asset keeps its identifier, per-frame features, semantic and source information, roles (such as internal or recovery-only), tracker hints, and quality metadata. For source rate $f$ and target rate $f'$, endpoint-preserving resampling uses
\begin{equation}
\begin{split}
N'&=\max\{1,\operatorname{round}((N-1)f'/f)+1\},\\
u_j&=j(N-1)/(N'-1).
\end{split}
\end{equation}
Position and joint angles are linearly interpolated and rotations use SLERP; the single-frame case is handled separately. The canonical asset rate is 30 Hz. Tracker adapters can resample again to their native reference rates. A successful conversion requires finite root/joint/time arrays, correct joint order and shape, and successful loading. It does not imply physical execution success.

For qpos-derived assets, velocities are forward differences with the last value repeated, followed by a 19-frame uniform filter with zero padding. Root angular velocity is obtained from $f\operatorname{Log}(R_{t+1}R_t^{-1})$. Recorded RL velocities may be retained, so not every asset uses this derivative/filter construction. The entry matcher uses its own central differences with one-sided endpoint differences.

\paragraph{Contact and windows.}
With local body points, a foot is marked in contact when its height above its within-clip minimum is at most $0.05$ m and its absolute vertical speed is at most $0.30$ m/s. Contact confidence combines the normalized height and speed margins with equal weights and is clipped to $[0,1]$. Missing body points give unknown contact with zero confidence. A stable frame satisfies $\|v\|_2\leq0.25$ m/s and $\|\dot\theta\|_2\leq0.90$ rad/s. A stable prefix or suffix must contain at least five frames; otherwise the corresponding first/last ten frames, or the whole shorter asset, form a low-confidence fallback. Metadata windows are half-open. The transition planner uses up to 60 head/tail frames when metadata windows are absent.

\paragraph{Entry proposal.}
A preliminary matcher searches the inclusive interval from the default start through $\min(N-2,\text{start}+60)$, so there can be 61 candidates. Its cost is
\begin{equation}
\begin{split}
C_{\rm entry}(t)={}&\operatorname{MAE}(\theta_x,\theta_t)+0.3d_{SO(3)}(r_x,r_t)\\
&+0.2\|v_x-\widetilde v_t\|_2+0.8|p_{x,z}-p_{t,z}|,
\end{split}
\end{equation}
where $d_{SO(3)}(r,s)=2\arccos|\langle\bar r,\bar s\rangle|$ for normalized quaternions. The joint-velocity weight is zero by default. Missing state terms are omitted. The near-optimal margin is zero, so exact ties select the earliest frame. Asset contracts can change weights and phase intervals, and compiled hints or explicit source intervals can bypass this matcher. The matcher only proposes an entry phase, which joins the candidates of entry matching; the continuity score of \cref{eq:boundary-search} makes the final choice.

\begin{wraptable}{r}{0.5\textwidth}
\centering
\vspace{-8pt}
\caption{The eight terms of the boundary error $\mathbf e$, with the weights $\mathbf w$ and thresholds $\boldsymbol\tau$ of \cref{eq:boundary-search}. Thresholds test the selected minimum-score candidate componentwise; weights act on raw units.}
\label{tab:boundary-params}
\footnotesize
\begin{tabular}{lrrl}
\toprule
Error component & Weight & Threshold & Unit\\
\midrule
Joint pose MAE & 1.00 & 0.18 & rad\\
Joint velocity MAE & 0.35 & 0.25 & rad/s\\
Relative root XY & 0.25 & 0.50 & m\\
Root height & 0.80 & 0.18 & m\\
Root up-vector angle & 0.50 & 0.25 & rad\\
Root linear velocity & 0.35 & 0.40 & m/s\\
Root angular velocity & 0.25 & 0.50 & rad/s\\
Known contact mismatch & 1.25 & 0 & binary\\
\bottomrule
\end{tabular}
\vspace{-6pt}
\end{wraptable}
\paragraph{Boundary error and entry matching.}
\Cref{tab:boundary-params} lists the eight terms of the boundary error $\mathbf e(b,b')$ between a source state $b$ and a target state $b'$. The joint terms are mean absolute differences over joints, the relative root XY term is the displacement from the window anchor in world axes, the up-vector term is the angle between the two root up-vectors, and the binary contact term counts only feet with known contact in both states. All terms except root XY, including the world-frame velocities, are evaluated before re-anchoring, and ties keep the earliest candidate frame. Entry matching minimizes first and then checks the thresholds, so an incompatible minimum is not replaced by a costlier compatible candidate. When it scores an entry frame, the frame's XY position is replaced by the measured one and its height is kept.

\begin{wraptable}{r}{0.45\textwidth}
\centering
\vspace{-8pt}
\caption{Bridge action pool. Bridges are existing assets, not learned transition policies; each must pass the pool checks and load at runtime.}
\label{tab:bridge-pool}
\footnotesize
\begin{tabular}{l}
\toprule
Action identifier\\
\midrule
\texttt{stable\_stand\_bridge}\\
\texttt{crouchwalk\_bridge}\\
\texttt{rl\_stabilize\_stand}\\
\texttt{idle\_stand\_sway}\\
\texttt{at\_ease\_then\_step\_back}\\
\texttt{step\_back\_right\_face\_right\_diagonal}\\
\texttt{fighting\_ready\_stance}\\
\texttt{jolt\_then\_stand\_tall}\\
\bottomrule
\end{tabular}
\vspace{-6pt}
\end{wraptable}
\paragraph{Bridge library.}
\Cref{tab:bridge-pool} lists the eight bridge actions. Bridge search ranks each candidate bridge $h$ by $C_{\rm in}(h)+C_{\rm out}(h)+0.05\,N_h/f_h$, the continuity scores of entering and leaving $h$ plus a penalty on its duration in seconds. The pool is manually assembled from existing motions. Candidate actions must be registered and loadable, and they must have a validated source asset (\cref{sec:appendix-routing}) that also passes a standing-transition check. Source, target, and previously used bridges are excluded. Candidate-pool quality factors and declared heights do not enter the score. Matching uses the loaded asset's entry/exit windows, and the duration penalty uses the full asset. A candidate is selected only if its entry frame precedes its exit frame, and the selected body slice includes both endpoints. If later bridge selection changes the preceding exit so that entry no longer precedes exit, the chain is stopped. There is no backtracking. The standing fallback loads the standing bridge directly and bypasses this pool filter.

The adaptive cutoff of \cref{eq:adaptive} is $\bar c=0.70$ if bridge search succeeds or the joint-pose term $e_\theta$ is at least $0.40$, and $\bar c=0.60$ otherwise. The boundary weights, thresholds, and adaptive cutoffs are fixed defaults set without held-out tuning.

\paragraph{Re-anchoring and interpolation.}
After each segment (an asset, a bridge, or an interpolation), the runtime reads the measured state $x$ and re-anchors the successor at its entry frame $j^\star$ to the robot's position and heading, with heading offset $\delta=\psi_x-\psi_{j^\star}$:
\begin{equation}
p'_t=R_z(\delta)(p_t-p_{j^\star})+p_x,\quad
r'_t=R_z(\delta)\,r_t,\quad \theta'_t=\theta_t,
\label{eq:reanchor}
\end{equation}
and rotates its root velocities by the same yaw. For a root-position or joint-angle coordinate $y$, let $F=24$, $T=(F-1)/f$, and $\sigma=j/(F-1)$ for frame $j=0,\dots,F-1$. The cubic Hermite transition is
\begin{equation}
\begin{split}
y(\sigma)={}&(2\sigma^3-3\sigma^2+1)y_0+(\sigma^3-2\sigma^2+\sigma)T\dot y_0\\
&+(-2\sigma^3+3\sigma^2)y_1+(\sigma^3-\sigma^2)T\dot y_1.
\end{split}
\label{eq:hermite}
\end{equation}
Endpoint velocities come from the measured state and the target reference when available; the endpoint span $T$ is $23/30$~s, while the 24 frames last $24/30$~s. Matching endpoint positions and velocities does not make the whole-body reference $C^2$ continuous or dynamically feasible.

For the standing fallback, $y(t)=\sum_{i=0}^{5}c_it^i$ applies to root position and named joint angles, with endpoint acceleration set to zero. More generally, define
\begin{equation}
\begin{aligned}
c_0&=y_0,\quad c_1=\dot y_0,\quad c_2=\ddot y_0/2,\\
P&=y_1-c_0-c_1T-c_2T^2,\\
V&=\dot y_1-c_1-2c_2T,\qquad A=\ddot y_1-2c_2,\\
c_3&=10P/T^3-4V/T^2+A/(2T),\\
c_4&=-15P/T^4+7V/T^3-A/T^2,\\
c_5&=6P/T^5-3V/T^4+A/(2T^3).
\end{aligned}
\end{equation}
At least three frames are required. For orientation, spherical de Casteljau interpolation uses $r_0$, $\operatorname{Exp}(T\omega_0/3)r_0$, $\operatorname{Exp}(-T\omega_1/3)r_1$, and $r_1$, with three, then two, then one short-arc SLERPs. Quaternion signs are made continuous; dot products above $0.9995$ use normalized linear interpolation. This rotational construction is not an angular-acceleration-matching quintic.

\paragraph{Runtime settings.}
Initialization enables a standing reset with 30 blend frames and 15 hold frames, so it is not a continuous transition from an arbitrary starting state. Ordinary recovery requires a stabilization/recovery asset with a safe reentry role, root height at least $0.35$ m, maximum absolute roll/pitch at most $0.6$ rad, and root speed at most $1.5$ m/s. Budgets are one attempt per segment and two per sequence. Recovery clears the preceding-action state and rematches the next action, without redoing a failed semantic action. No default get-up asset is configured. Tracker future-reference preview, where consumed by an adapter, provides context rather than an MPC rollout search.

\section{Language Planning}
\label{sec:appendix-language}
\paragraph{Planner and catalog.}
A frozen external LLM maps the natural-language instruction to an ordered list of canonical action identifiers. The configured language-model identifier is \texttt{deepseek-v4-flash}, with temperature zero, at most four attempts, a 60 s request timeout, and an initial 1,024-token output budget that can double to 8,192 after an empty, truncated, or unparsable response. The configuration does not pin an immutable service version. The catalog lists an action only if one of its assets is validated (\cref{sec:appendix-routing}) and passes a standing-transition check, and it hides internal and recovery-only actions. The LLM output is schema-checked and matched to catalog actions: retrieval uses lowercase ASCII alphanumeric tokens, treating underscores and hyphens as separators; exact identifiers score 1, aliases 0.9, and other matches use positive Jaccard overlap. Equal scores retain catalog order. The planner returns action identifiers rather than trajectories, and the runtime resolves each action to a validated asset (\cref{sec:representation}). The LLM receives no robot sensor state; the runtime consumes the measured state. Missing-action decomposition is an optional bounded repair path and does not guarantee that every replacement is executable.

\paragraph{Planning in the experiments.}
The planner receives the natural-language prompt and the action catalog, never the ground-truth action labels. Its predicted action order and resolved assets are fixed once and shared by all methods. On the benchmark, it recovers the ground-truth sequence of canonical action identifiers for all 400 instructions, so planning errors do not affect any comparison. Object-related labels name reference motions; they do not imply manipulation success.

\section{Validation Evidence Details}
\label{sec:appendix-routing}
\paragraph{Validation rollouts.}
Each of 1,010 of the 3,102 library assets is executed alone with each candidate tracker in three independent validation rollouts, 24,240 in total. A rollout succeeds under the trial criterion of the experiments (\cref{sec:setup}): the asset completes without a fall, timeout, or abnormal termination, and $E_{\max}\leq0.3$~m. Validation thus applies the criterion of the sequence trials to one asset at a time. The validation MPJPE $\epsilon_{ak}$ (\cref{sec:routing}) and velocity error $\nu_{ak}$ average the MPJPE and $E_v$ of \cref{sec:metrics} over the three rollouts, failed rollouts included; only non-finite values are dropped.

\paragraph{Aggregation and ties.}
Because every candidate has three rollouts per asset, the pooled success $\hat p_k$ equals the mean per-asset success rate, so comparing it amounts to comparing total successes. Errors are averaged over assets with equal weight, and no instruction contains the same asset twice. All criteria are compared exactly. Ties left by \cref{eq:assignment} are broken by the mean velocity error $\bar\nu_k$ and then by ascending tracker identifier, which orders the candidates as FRoM-W1 (whose H-ACT identifier sorts first), HoloMotion, Humanoid-GPT, MimicLite, OpenTrack, ProtoMotions GTP, SONIC, and TWIST2.

\paragraph{Deciding criterion.}
Pooled validation success decides the assignment of 79 of the 400 benchmark instructions and MPJPE that of 302; 18 instructions have a single eligible tracker, and one has none. Velocity error and the tracker identifier never decide. Validation success saturates because 377 of the 399 assignable instructions have at least one candidate that succeeds in every validation rollout of every asset.

\paragraph{Native profiles.}
\Cref{tab:native} lists the eight candidate profiles; all use 29 reference and 29 controlled joints. Observation histories, models, PD gains, and action scales are those of each native controller.
\begin{table*}[tbp]
\centering
\caption{Candidate tracker profiles. All use 29 reference and 29 controlled joints. Rates are control/reference/physics in Hz. Native adapters map policy outputs to the 29 joints (FRoM-W1 maps 21 outputs); ``auto'' denotes inputs defined by model metadata.}
\label{tab:native}
\footnotesize
\begin{tabular}{lrrr}
\toprule
Tracker & Observation dimension & Policy output dimension & Rates (Hz)\\
\midrule
SONIC & 1,762 & 29 & 50/50/500\\
MimicLite & 839 & 29 & 50/50/200\\
TWIST2 & 1,432 & 29 & 100/50/1,000\\
OpenTrack & 162 & 29 & 50/50/500\\
FRoM-W1 & 1,821 & 21 & 50/50/1,000\\
HoloMotion & 604 & 29 & 50/50/500\\
ProtoMotions GTP & auto & 29 & 50/50/1,000\\
Humanoid-GPT & 136 & 29 & 50/50/250\\
\bottomrule
\end{tabular}
\end{table*}

\section{Benchmark and Metric Definitions}
\label{sec:metrics}
The benchmark lists 400 instructions in four length groups of 100; each instruction has exactly the stated number of ordered actions. Together they contain 1,800 action slots and 100 distinct action identifiers. Ground-truth action labels are used only for evaluation. \Cref{tab:trial-counts} gives the resulting trial and boundary counts per method.
\begin{table}[tbp]
\centering
\caption{Benchmark composition with trial and semantic-boundary counts per method. Each instruction has five test seeds.}
\label{tab:trial-counts}
\footnotesize
\begin{tabular}{rrrrr}
\toprule
Length & Instructions & Action slots & Trials & Boundaries\\
\midrule
2 & 100 & 200 & 500 & 500\\
3 & 100 & 300 & 500 & 1,000\\
5 & 100 & 500 & 500 & 2,000\\
8 & 100 & 800 & 500 & 3,500\\
\midrule
All & 400 & 1,800 & 2,000 & 7,000\\
\bottomrule
\end{tabular}
\end{table}

\paragraph{Root-relative error.}
For execution and reference trajectories, let $r(t,j)=p(t,j)-p(t,\mathrm{root})$ for the same named body link $j$. No heading alignment is applied. For $J=23$ common non-root links and $T$ observed aligned frames,
\begin{equation}
 e_t=\frac{1}{J}\sum_{j=1}^{J}\|r_{\rm exec}(t,j)-r_{\rm ref}(t,j)\|_2,
 \quad \mathrm{MPJPE}=\frac{1}{T}\sum_{t=1}^{T}e_t.
\end{equation}
The maximum-frame criterion is $\max_t e_t\leq0.3$ m, not maximum single-link distance. Completeness and the absence of a fall, timeout, or abnormal termination are separate necessary conditions, so a short successful-looking prefix does not pass a failed sequence.

\paragraph{Velocity and discontinuity.}
Relative velocity is $v(t,j)=[r(t+1,j)-r(t,j)]/\Delta t$, and
\begin{equation}
E_v=\frac{1}{(T-1)J}\sum_{t=1}^{T-1}\sum_j\|v_{\rm exec}(t,j)-v_{\rm ref}(t,j)\|_2.
\end{equation}
Sequence, action-interior, and transition-window errors are computed separately. For $D$ evaluated joint angles and adjacent samples straddling boundary $b$,
\begin{equation}
J_{\theta}=\frac{1}{D}\sum_j|\theta_j(t_b^+)-\theta_j(t_b^-)|,\quad
J_{\dot\theta}=\frac{1}{D}\sum_j|\dot\theta_j(t_b^+)-\dot\theta_j(t_b^-)|.
\end{equation}
Both reference and executed trajectories are measured. Peak velocity, mean acceleration, and peak acceleration of the executed joints in a fixed window around each boundary complement these boundary jumps.

\paragraph{Denominators and aggregation.}
SSR and FR divide by all trials. TSR divides by attempted boundaries, including rejections. Unreached boundaries are counted separately and contribute no error samples. The 7,000 semantic boundaries per transition method exclude the boundaries inside bridge chains. Mean errors are computed per trial and then averaged across trials; because every instruction has five seeds, this equals instruction-level averaging. Observed prefixes of failed trials are retained, and \cref{tab:transition-failures} reports the observed frames. Errors are therefore trial-weighted rather than frame-weighted.

Paired comparisons preserve instruction and seed identity. The five seeds of an instruction are correlated observations, so the confidence intervals and $p$ values of Experiments 1 and 2 resample whole instructions; Experiments 3 and 4, the ablation and robustness comparisons, use McNemar tests on paired trials. Holm correction is applied within each family of comparisons against CHOREO, one family per experiment. Runtime errors and rejections remain in the denominators.

\paragraph{Falls.}
A fall is recorded when, in any frame, the root height drops below $0.20$ m or the torso tilt exceeds $120^\circ$, or when a tracker's native termination condition fires.

\section{Single-Asset Execution}
\label{sec:single-asset}
\begin{table*}[tbp]
\centering
\caption{Single-asset execution. Each candidate tracker executes every one of the 3,102 library assets alone, without transitions, with five seeds (15,510 rollouts per tracker). ESR and FR divide by all rollouts. Errors include the observed prefixes of failed rollouts and weight assets equally; MPJPE and global MPJPE are in m, $E_v$ in m/s. $\uparrow$/$\downarrow$: higher/lower is better; bold/underline: best/second-best per column.}
\label{tab:single-overall}
\footnotesize
\begin{tabular}{lrcrcccc}
\toprule
& & & & & \multicolumn{3}{c}{Error $\downarrow$}\\
\cmidrule(lr){6-8}
Tracker & Successes & ESR (\%) $\uparrow$ & Falls & FR (\%) $\downarrow$ & MPJPE & $E_v$ & Global MPJPE\\
\midrule
SONIC & \underline{15,398} & \underline{99.28} & 112 & \phantom{0}0.72 & \underline{0.0252} & \textbf{0.129} & 0.634\\
MimicLite & 14,916 & 96.17 & 578 & \phantom{0}3.73 & \textbf{0.0248} & 0.166 & 0.281\\
TWIST2 & 4,312 & 27.80 & 3,982 & 25.67 & 0.0889 & 0.314 & 0.391\\
OpenTrack & 6,220 & 40.10 & 8,793 & 56.69 & 0.2079 & 0.413 & 1.237\\
FRoM-W1 & 14,439 & 93.09 & 267 & \phantom{0}1.72 & 0.1532 & 0.307 & \underline{0.218}\\
HoloMotion & \textbf{15,442} & \textbf{99.56} & \underline{63} & \phantom{0}\underline{0.41} & 0.0266 & \underline{0.144} & \textbf{0.075}\\
ProtoMotions GTP & 14,281 & 92.08 & 1,193 & \phantom{0}7.69 & 0.0399 & 0.202 & 0.586\\
Humanoid-GPT & 14,556 & 93.85 & \textbf{1} & \phantom{0}\textbf{0.01} & 0.0305 & 0.173 & 0.257\\
\bottomrule
\end{tabular}
\end{table*}
\begin{table*}[tbp]
\centering
\caption{Single-asset execution by asset source. Columns and units are those of \cref{tab:single-overall}; within each source, assets are weighted equally.}
\label{tab:single-source}
\footnotesize
\begin{tabular}{lrcrcccc}
\toprule
& & & & & \multicolumn{3}{c}{Error $\downarrow$}\\
\cmidrule(lr){6-8}
Tracker & Successes & ESR (\%) $\uparrow$ & Falls & FR (\%) $\downarrow$ & MPJPE & $E_v$ & Global MPJPE\\
\midrule
\multicolumn{8}{l}{\textit{Motion data: 2,950 assets, 14,750 rollouts per tracker}}\\
SONIC & 14,638 & \phantom{0}99.24 & 112 & \phantom{0}0.76 & 0.0256 & 0.132 & 0.659\\
MimicLite & 14,161 & \phantom{0}96.01 & 575 & \phantom{0}3.90 & 0.0255 & 0.171 & 0.291\\
TWIST2 & 3,637 & \phantom{0}24.66 & 3,972 & 26.93 & 0.0924 & 0.323 & 0.402\\
OpenTrack & 5,539 & \phantom{0}37.55 & 8,721 & 59.13 & 0.2149 & 0.424 & 1.268\\
FRoM-W1 & 13,679 & \phantom{0}92.74 & 267 & \phantom{0}1.81 & 0.1538 & 0.312 & 0.220\\
HoloMotion & 14,682 & \phantom{0}99.54 & 63 & \phantom{0}0.43 & 0.0272 & 0.149 & 0.078\\
ProtoMotions GTP & 13,536 & \phantom{0}91.77 & 1,178 & \phantom{0}7.99 & 0.0406 & 0.206 & 0.596\\
Humanoid-GPT & 13,796 & \phantom{0}93.53 & 1 & \phantom{0}0.01 & 0.0311 & 0.177 & 0.264\\
\midrule
\multicolumn{8}{l}{\textit{RL rollouts: 76 assets, 380 rollouts per tracker}}\\
SONIC & 380 & 100.00 & 0 & \phantom{0}0.00 & 0.0156 & 0.098 & 0.250\\
MimicLite & 380 & 100.00 & 0 & \phantom{0}0.00 & 0.0101 & 0.084 & 0.073\\
TWIST2 & 355 & \phantom{0}93.42 & 0 & \phantom{0}0.00 & 0.0226 & 0.225 & 0.247\\
OpenTrack & 375 & \phantom{0}98.68 & 5 & \phantom{0}1.32 & 0.0565 & 0.247 & 0.851\\
FRoM-W1 & 380 & 100.00 & 0 & \phantom{0}0.00 & 0.1523 & 0.291 & 0.177\\
HoloMotion & 380 & 100.00 & 0 & \phantom{0}0.00 & 0.0151 & 0.095 & 0.029\\
ProtoMotions GTP & 380 & 100.00 & 0 & \phantom{0}0.00 & 0.0254 & 0.159 & 0.624\\
Humanoid-GPT & 380 & 100.00 & 0 & \phantom{0}0.00 & 0.0150 & 0.111 & 0.139\\
\midrule
\multicolumn{8}{l}{\textit{Generated motions: 76 assets, 380 rollouts per tracker}}\\
SONIC & 380 & 100.00 & 0 & \phantom{0}0.00 & 0.0167 & 0.039 & 0.066\\
MimicLite & 375 & \phantom{0}98.68 & 3 & \phantom{0}0.79 & 0.0147 & 0.050 & 0.096\\
TWIST2 & 320 & \phantom{0}84.21 & 10 & \phantom{0}2.63 & 0.0194 & 0.077 & 0.091\\
OpenTrack & 306 & \phantom{0}80.53 & 67 & 17.63 & 0.0870 & 0.155 & 0.395\\
FRoM-W1 & 380 & 100.00 & 0 & \phantom{0}0.00 & 0.1299 & 0.131 & 0.157\\
HoloMotion & 380 & 100.00 & 0 & \phantom{0}0.00 & 0.0163 & 0.028 & 0.036\\
ProtoMotions GTP & 365 & \phantom{0}96.05 & 15 & \phantom{0}3.95 & 0.0294 & 0.088 & 0.187\\
Humanoid-GPT & 380 & 100.00 & 0 & \phantom{0}0.00 & 0.0217 & 0.048 & 0.105\\
\bottomrule
\end{tabular}
\end{table*}
This test checks that each candidate tracker is correctly integrated into our simulation environment and can execute the library assets on their own. It is separate from the validation of \cref{sec:routing} and covers the whole library. Each of the eight trackers executes each of the 3,102 assets (\cref{sec:setup}) alone, without transitions, with five seeds (11, 23, 37, 53, and 71). This gives 15,510 rollouts per tracker and 124,080 in total; \cref{tab:single-source} lists them by source. A rollout succeeds under the rule of the main experiments: the asset completes without a fall, timeout, or abnormal termination, and $E_{\max}\leq0.3$~m. The execution success rate (ESR) and FR divide by all rollouts. MPJPE and $E_v$ follow \cref{sec:metrics}; global MPJPE omits the root subtraction and therefore also captures root drift. Errors are averaged over the observed frames of each rollout, failed rollouts included, and then over rollouts, which weights assets equally because each has five seeds. Every tracker keeps its native settings (\cref{tab:native}), so the rows compare complete systems rather than policy weights.

\paragraph{Failure accounting.}
Conversion failures and execution failures are counted separately. Only assets that pass the conversion checks of \cref{sec:appendix-representation} enter the library and this test, so ESR and FR contain execution failures only. Each asset also runs alone, so no transition failure enters them either; ESR is a per-asset rate and is reported apart from the sequence-level SSR and TSR.

\paragraph{Results.}
Six of the eight trackers succeed in over 92\% of their rollouts, and HoloMotion and SONIC succeed in 99.56\% and 99.28\%, respectively (\cref{tab:single-overall}). Assets from every source are executable (\cref{tab:single-source}): six trackers succeed in all 380 rollouts on RL-rollout assets, four in all 380 on generated motions, and every tracker in over 80\% of both. The two weakest trackers, TWIST2 and OpenTrack, still succeed in 93.42\% and 98.68\% of the rollouts on RL-rollout assets, which indicates that their integration works; their low overall ESR (27.80\% and 40.10\%) comes from motion data (24.66\% and 37.55\%), consistent with their low SSR in \cref{tab:assignment}.
Each tracker has its lowest ESR and highest MPJPE and $E_v$ on motion data. No tracker leads every metric: HoloMotion has the highest ESR and lowest global MPJPE, MimicLite the lowest MPJPE, SONIC the lowest $E_v$, and Humanoid-GPT the fewest falls. Most failures of TWIST2, FRoM-W1, and Humanoid-GPT involve no fall, whereas falls cause at least 92\% of those of the other five trackers. As failed rollouts contribute only their observed prefixes, errors should be read with ESR.

\section{Extended Transition Evaluation}
\label{sec:transition-extra}
Main-paper \cref{tab:transition-ref,tab:transition-exec} report the constructed references and their execution. \Cref{tab:transition-counts,tab:transition-errors,tab:transition-boundaries,tab:transition-smoothness,tab:transition-paired,tab:transition-denominators,tab:transition-failures} complement these with trial counts, all tracking scopes, reference\slash executed discontinuity, window smoothness, paired inference, semantic-boundary denominators, and failure\slash coverage statistics. All comparisons use the same five methods and shared tracker assignments.

Each failed trial has one primary category: physical, transition\slash tracking, assignment, or resource\slash runtime\slash timeout. Falls are flagged as separate events, so event counts need not equal category counts. Error statistics should be read together with these categories and the observed-frame coverage, because a trial that fails early has low error over its short observed prefix. Paired SSR differences are CHOREO minus the indicated baseline, in pp. Of CHOREO's 79 falls, 72 follow a standing fallback, 4 a Hermite transition, and 3 occur before the first boundary; 54 occur inside a transition and 25 inside an action. By length, 55 occur at $L=8$, 18 at $L=5$, 6 at $L=3$, and none at $L=2$. The actions with the most falls are a spinning back kick (7), running backward with a turn (5), and a reverse spinning crescent kick (4).

\begin{table*}[tbp]
\centering
\caption{Transition comparison: overall success and stability. Each method has 2,000 trials; TSR divides by the attempted boundaries in \cref{tab:transition-denominators}. Bold and underline mark the best and second-best values in each column: highest successes, SSR, and TSR; lowest falls and FR.}
\label{tab:transition-counts}
\footnotesize
\begin{tabular}{lrrrrr}
\toprule
Method & Successes & SSR (\%) & TSR (\%) & Falls & FR (\%)\\
\midrule
Direct Concatenation & 1661 & 83.05 & \textbf{99.48} & 306 & 15.30\\
Fixed Hermite & 1734 & 86.70 & 98.86 & 251 & 12.55\\
Fixed Stand Bridge & \underline{1861} & \underline{93.05} & 98.95 & \underline{124} & \underline{6.20}\\
Motion Matching & 1742 & 87.10 & 97.00 & 249 & 12.45\\
CHOREO & \textbf{1916} & \textbf{95.80} & \underline{99.07} & \textbf{79} & \textbf{3.95}\\
\bottomrule
\end{tabular}
\end{table*}%
\begin{table*}[tbp]
\centering
\caption{Transition comparison: tracking errors with equal weight per trial, including the observed prefixes of failed trials. MPJPE is in m and velocity error $E_v$ in m/s. Whole-sequence, action-interior, and transition-window errors are computed separately. Bold and underline mark the lowest and second-lowest values in each column.}
\label{tab:transition-errors}
\footnotesize
\begin{tabular}{lrrrrr}
\toprule
Method & Sequence MPJPE & Transition MPJPE & Sequence $E_v$ & Interior $E_v$ & Transition $E_v$\\
\midrule
Direct Concatenation & 0.0406 & 0.0536 & 0.173 & 0.151 & 0.549\\
Fixed Hermite & 0.0346 & 0.0322 & 0.147 & 0.136 & 0.204\\
Fixed Stand Bridge & \textbf{0.0279} & \textbf{0.0245} & \textbf{0.121} & \textbf{0.125} & \textbf{0.121}\\
Motion Matching & 0.0313 & 0.0315 & 0.135 & 0.128 & 0.151\\
CHOREO & \underline{0.0280} & \underline{0.0251} & \underline{0.123} & \underline{0.126} & \underline{0.126}\\
\bottomrule
\end{tabular}
\end{table*}%
\begin{table*}[tbp]
\centering
\caption{Reference and executed boundary discontinuity. $J_\theta$ is mean angle jump (rad); $J_{\dot\theta}$ is mean velocity jump (rad/s). All are lower-is-better diagnostics. Bold and underline mark the lowest and second-lowest values in each column.}
\label{tab:transition-boundaries}
\footnotesize
\begin{tabular}{lrrrr}
\toprule
Method & Ref. $J_\theta$ & Ref. $J_{\dot\theta}$ & Exec. $J_\theta$ & Exec. $J_{\dot\theta}$\\
\midrule
Direct Concatenation & 0.3358 & 0.438 & 0.0430 & 2.683\\
Fixed Hermite & 0.0421 & 0.314 & 0.0183 & 1.106\\
Fixed Stand Bridge & \textbf{0.0325} & \textbf{0.212} & \underline{0.0164} & 0.969\\
Motion Matching & 0.0378 & 0.237 & \textbf{0.0160} & \underline{0.941}\\
CHOREO & \underline{0.0342} & \underline{0.216} & \textbf{0.0160} & \textbf{0.938}\\
\bottomrule
\end{tabular}
\end{table*}%
\begin{table*}[tbp]
\centering
\caption{Executed joint motion in a fixed window around each boundary. Lower values indicate less abrupt motion. Bold and underline mark the lowest and second-lowest values in each column.}
\label{tab:transition-smoothness}
\footnotesize
\begin{tabular}{lrrr}
\toprule
Method & Peak velocity (rad/s) & Mean acceleration (rad/s$^2$) & Peak acceleration (rad/s$^2$)\\
\midrule
Direct Concatenation & 11.57 & 44.4 & 556.6\\
Fixed Hermite & 5.15 & 19.8 & 277.5\\
Fixed Stand Bridge & 4.41 & \underline{15.1} & 233.6\\
Motion Matching & \underline{4.28} & 15.8 & \underline{231.5}\\
CHOREO & \textbf{4.22} & \textbf{14.5} & \textbf{222.0}\\
\bottomrule
\end{tabular}
\end{table*}%
\begin{table*}[tbp]
\centering
\caption{Paired SSR comparison, CHOREO minus baseline, in percentage points. Confidence intervals (95\%) and Holm-adjusted $p$ values come from paired resampling of whole instructions (\cref{sec:metrics}).}
\label{tab:transition-paired}
\footnotesize
\begin{tabular}{lrrrr}
\toprule
Baseline & Difference (pp) & 95\% CI & Holm $p$ & SSR advantage supported\\
\midrule
Direct Concatenation & 12.75 & [9.65, 16.05] & 0.0004 & Yes\\
Fixed Hermite & 9.10 & [6.50, 11.90] & 0.0004 & Yes\\
Fixed Stand Bridge & 2.75 & [1.05, 4.65] & 0.0029 & Yes\\
Motion Matching & 8.70 & [6.05, 11.35] & 0.0004 & Yes\\
\bottomrule
\end{tabular}
\end{table*}%
\begin{table*}[tbp]
\centering
\caption{Semantic boundaries per method (7,000 planned). Attempted includes rejections; unreached boundaries are counted separately. Boundaries inside inserted bridge chains are not counted.}
\label{tab:transition-denominators}
\footnotesize
\begin{tabular}{lrrrrr}
\toprule
Method & Planned & Attempted & Successful & Rejected & Unreached\\
\midrule
Direct Concatenation & 7,000 & 6,307 & 6,274 & 0 & 693\\
Fixed Hermite & 7,000 & 6,420 & 6,347 & 0 & 580\\
Fixed Stand Bridge & 7,000 & 6,670 & 6,600 & 0 & 330\\
Motion Matching & 7,000 & 6,469 & 6,275 & 0 & 531\\
CHOREO & 7,000 & 6,743 & 6,680 & 0 & 257\\
\bottomrule
\end{tabular}
\end{table*}%
\begin{table*}[tbp]
\centering
\caption{Failure categories and observation coverage for the transition comparison. Each failed trial has one primary category. Assignment failures are the five trials of the one instruction for which no tracker is eligible for all of its assets; all methods share them. Bold and underline mark the lowest and second-lowest counts in the Physical and Transition/tracking columns.}
\label{tab:transition-failures}
\footnotesize
\begin{tabular}{lrrrrr}
\toprule
Method & Physical & Transition/tracking & Assignment & Resource/runtime/timeout & Observed frames\\
\midrule
Direct Concatenation & 306 & 28 & 5 & 0 & 2,839,664\\
Fixed Hermite & 251 & 10 & 5 & 0 & 3,160,214\\
Fixed Stand Bridge & \underline{124} & 10 & 5 & 0 & 4,899,588\\
Motion Matching & 249 & \underline{4} & 5 & 0 & 4,273,928\\
CHOREO & \textbf{79} & \textbf{0} & 5 & 0 & 3,905,672\\
\bottomrule
\end{tabular}
\end{table*}%
\begin{table*}[tbp]
\centering
\caption{Instructions of the real-robot experiment (\cref{sec:exp-hardware}). All nine are $L=5$ instructions that succeed in simulation; each is executed five times, zero-shot, on the physical Unitree G1. Hard instructions contain several dynamic actions or an action with a large whole-body posture change. Actions are the benchmark's canonical identifiers.}
\label{tab:hardware}
\footnotesize
\begin{tabular}{lp{0.56\textwidth}cc}
\toprule
Instruction ID & Action sequence & Difficulty & Success\\
\midrule
\texttt{sequence\_5\_000\_v0} & \raggedright \texttt{stand\_ready} $\to$ \texttt{stand\_to\_walk} $\to$ \texttt{conversation\_gestures} $\to$ \texttt{walk\_turn\_left\_135} $\to$ \texttt{at\_ease\_then\_step\_back} & Easy & 5/5\\
\texttt{sequence\_5\_015\_v0} & \raggedright \texttt{stand\_still} $\to$ \texttt{pick\_up\_box} $\to$ \texttt{wave\_right\_hand} $\to$ \texttt{lift\_box} $\to$ \texttt{at\_ease\_then\_step\_back} & Easy & 5/5\\
\texttt{sequence\_5\_035\_v0} & \raggedright \texttt{at\_ease\_then\_step\_back} $\to$ \texttt{throw\_hard} $\to$ \texttt{raise\_and\_spread\_both\_arms} $\to$ \texttt{throw\_and\_catch} $\to$ \texttt{crouch\_down} & Easy & 5/5\\
\texttt{sequence\_5\_075\_v0} & \raggedright \texttt{at\_ease\_then\_step\_back} $\to$ \texttt{throw\_hard} $\to$ \texttt{wave\_right\_hand} $\to$ \texttt{throw\_and\_catch} $\to$ \texttt{fighting\_ready\_stance} & Easy & 5/5\\
\texttt{sequence\_5\_084\_v0} & \raggedright \texttt{fighting\_ready\_stance} $\to$ \texttt{waltz} $\to$ \texttt{lean\_forward\_arms\_open} $\to$ \texttt{dance} $\to$ \texttt{sway\_body} & Easy & 5/5\\
\texttt{sequence\_5\_085\_v0} & \raggedright \texttt{at\_ease\_then\_step\_back} $\to$ \texttt{walk\_to\_pick\_up\_box} $\to$ \texttt{wave\_right\_hand} $\to$ \texttt{lift\_box} $\to$ \texttt{horse\_squat} & Easy & 5/5\\
\midrule
\texttt{sequence\_5\_008\_v0} & \raggedright \texttt{crouch\_down} $\to$ \texttt{push\_kick\_right} $\to$ \texttt{look\_around} $\to$ \texttt{roundhouse\_left} $\to$ \texttt{stand\_still} & Hard & 5/5\\
\texttt{sequence\_5\_057\_v0} & \raggedright \texttt{stand\_still} $\to$ \texttt{step\_forward\_once} $\to$ \texttt{axe\_kick} $\to$ \texttt{walk\_to\_stand} $\to$ \texttt{stand\_ready} & Hard & 5/5\\
\texttt{sequence\_5\_076\_v0} & \raggedright \texttt{jump\_in\_place} $\to$ \texttt{shy\_body\_sway} $\to$ \texttt{axe\_kick} $\to$ \texttt{turn\_right\_90} $\to$ \texttt{horse\_squat} & Hard & 5/5\\
\midrule
All & & & 45/45\\
\bottomrule
\end{tabular}
\end{table*}
\section{Ablation and Robustness}
\label{sec:ablation-appendix}
\paragraph{Ablated variants.}
Each variant in \cref{fig:ablation} removes one component of the composition rule (\cref{sec:composition}) and keeps CHOREO's tracker assignment, assets, and seeds. The w/o measured state variant plans every boundary, Hermite start point, and re-anchoring from the nominal endpoint of the previous asset, which amounts to splicing the whole reference before execution. The w/o bridge search variant keeps only Hermite transitions and the standing fallback. The w/o adaptive decision variant drops the condition $c^\star<\bar c$ from \cref{eq:adaptive}, so only compatible boundaries receive a Hermite transition. The w/o standing fallback variant uses a Hermite transition wherever \cref{eq:adaptive} would choose \textsf{Stand}. The w/o entry matching variant enters every asset at its first frame instead of a searched entry frame. The w/o continuity score variant drops the continuity score, the weighted error of \cref{eq:boundary-search}, from boundary selection.

\paragraph{Perturbations.}
A horizontal velocity impulse of 0.3, 0.6, or 0.9~m/s is applied to the robot's root before boundaries; magnitude zero is the unperturbed condition. Every method keeps CHOREO's tracker assignment, assets, and seeds, giving 2,000 trials per method and magnitude. \Cref{fig:robustness} compares CHOREO with the w/o measured state variant, Fixed Stand Bridge, and Fixed Hermite. Without impulses, each method reaches the same SSR as in \cref{tab:transition-exec} or \cref{fig:ablation}.

\section{Composition Across Asset Sources}
\label{sec:cross-source}
We test whether CHOREO composes assets from different sources. Seven actions use assets that are not motion data: six use RL rollouts and one uses a generated motion (\cref{tab:cross-source}). \Cref{fig:cross-source} groups the boundaries that CHOREO attempts by the sources of the two assets. CHOREO succeeds at every boundary between assets from different sources. The 766 boundaries between a motion-data asset and an RL rollout or a generated motion, in either order, cover 149 distinct action pairs and 70 different motion-data actions. The five boundaries between RL rollouts of two different policies and all 544 executed segments of the seven actions also succeed. Boundaries between two motion-data assets succeed in 98.9\% of 5,972 attempts, so every boundary that CHOREO fails joins two motion-data assets.
\begin{table}[tbp]
\begin{minipage}[t]{0.49\textwidth}
\vspace{0pt}
\centering
\includegraphics[width=\linewidth]{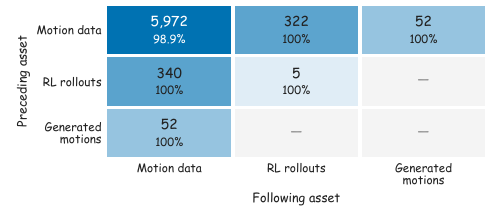}
\captionof{figure}{Boundaries that CHOREO attempts, by the source of the asset before (rows) and after (columns) each boundary. A cell gives the number of boundaries, shaded on a log scale, and the percentage that succeed. A dash marks source pairs that no boundary joins.}
\label{fig:cross-source}
\end{minipage}\hfill
\begin{minipage}[t]{0.48\textwidth}
\vspace{0pt}
\centering
\caption{Actions whose assets are RL rollouts or generated motions. Instr.: instructions containing the action; Segments: executions of the action; In and Out: boundaries from and to a motion-data asset. Every segment and boundary listed succeeds, and the All row counts each instruction once.}
\label{tab:cross-source}
\footnotesize
\begin{tabular}{lrrrr}
\toprule
Action & Instr. & Segments & In & Out\\
\midrule
\multicolumn{5}{l}{\textit{RL rollouts}}\\
\texttt{horse\_squat} & 54 & 253 & 143 & 121\\
\texttt{turn\_right\_180} & 11 & 50 & 45 & 45\\
\texttt{walk\_forward} & 10 & 50 & 40 & 45\\
\texttt{turn\_right\_45} & 10 & 50 & 30 & 50\\
\texttt{turn\_left\_45} & 10 & 49 & 44 & 44\\
\texttt{turn\_left\_90} & 7 & 35 & 20 & 35\\
\midrule
\multicolumn{5}{l}{\textit{Generated motions}}\\
\texttt{wave\_right\_hand} & 12 & 57 & 52 & 52\\
\midrule
All & 103 & 544 & 374 & 392\\
\bottomrule
\end{tabular}
\end{minipage}
\end{table}

\section{Real-Robot Deployment}
\label{sec:hardware-appendix}
\Cref{tab:hardware} lists the nine instructions of Experiment~5 (\cref{sec:exp-hardware}) with their difficulty. Each succeeds in all five of its trials.

\section{Tracker Assignment Analyses}
\label{sec:assignment-analyses}
\enlargethispage{\baselineskip}
\begin{wraptable}{r}{0.46\textwidth}
\vspace{-8pt}
\centering
\caption{The 302 instructions on which several trackers tie for the highest validation success, with their tied trackers ranked by validation MPJPE; rank 1 is CHOREO's choice. Every tied tracker runs the instruction's five seeds under CHOREO's composition; rank $r$ covers the instructions with at least $r$ tied trackers.}
\label{tab:tie-rank}
\footnotesize
\begin{tabular}{rrrr}
\toprule
Rank & Instructions & Successes / trials & SSR (\%)\\
\midrule
1 & 302 & 1,482 / 1,510 & 98.15\\
2 & 302 & 1,330 / 1,510 & 88.08\\
3 & 247 & 1,101 / 1,235 & 89.15\\
4 & 211 & 940 / 1,055 & 89.10\\
5 & 172 & 724 / 860 & 84.19\\
6 & 113 & 487 / 565 & 86.19\\
7 & 55 & 237 / 275 & 86.18\\
8 & 20 & 95 / 100 & 95.00\\
\bottomrule
\end{tabular}
\vspace{-6pt}
\end{wraptable}
\paragraph{Precision among tied trackers.}
On the 302 instructions decided by MPJPE, every tied tracker was executed with the instruction's five seeds under CHOREO's composition. Ranked by validation MPJPE (\cref{tab:tie-rank}), the most precise tracker, which CHOREO chooses, succeeds in 98.15\% of its trials, whereas every other rank stays between 84.19\% and 95.00\% (rank 8 covers only 20 instructions). Paired over instructions and seeds, rank 1 exceeds rank 2 by 10.07~pp (95\% CI [6.69, 13.77], resampling instructions), and by 5.06, 5.43, 14.10, and 18.93~pp at $L=2,3,5,8$. Over the 1,422 instruction--tracker pairs, validation MPJPE correlates with sequence success (Spearman $\rho=-0.32$), more strongly for longer instructions ($-0.29$ at $L=2$, $-0.40$ at $L=8$). Among boundaries whose entry is reached, the failure rate after the boundary grows with the entry error for the other tied trackers but stays at or below 0.31\% for the most precise one (\cref{tab:tie-entry}); falls inside a transition, before the entry is reached, have no entry error and are not covered by this comparison.

\paragraph{Tracker-pool size.}
Applying the rule of \cref{eq:assignment} within every non-empty subset of the eight trackers (255 subsets), with trial outcomes taken from the runs above, reproduces CHOREO's 1,916 successes for the full pool. Averaged over subsets of the same size, SSR rises with every added tracker, from 46.84\% for one to 95.80\% for all eight, whereas uniform random choice among the eligible trackers levels off near 78\% (\cref{tab:pool-size}). A one-tracker subset counts as failures the instructions for which its tracker is not eligible, so its values differ from the fixed-tracker rows of \cref{tab:assignment}. The best subsets in hindsight, starting with SONIC and HoloMotion, reach 96.50\%, 0.70~pp above the full pool. Removing SONIC or HoloMotion from the full pool lowers SSR to 85.80\% or 86.70\%, and removing Humanoid-GPT to 95.55\%; removing MimicLite, which tracks single assets most precisely (\cref{tab:single-overall}), raises it to 96.50\%, and removing any other tracker leaves it unchanged.

\begin{table}[!hb]
\begin{minipage}[t]{0.52\textwidth}
\vspace{0pt}
\centering
\caption{Failures after a boundary among the tied trackers, by quartile of entry error: the root-relative position error $e_t$ (\cref{sec:metrics}) between the measured state and the selected entry frame. Only boundaries whose entry is reached have an entry error.}
\label{tab:tie-entry}
\footnotesize
\setlength{\tabcolsep}{5.7pt}
\begin{tabular}{lrr}
\toprule
Entry error (m) & Rank 1 & Other ranks\\
\midrule
$<0.0184$ & 0 / 1,888 (0.00\%) & 7 / 2,594 (0.27\%)\\
0.0184--0.0265 & 0 / 1,185 (0.00\%) & 20 / 3,296 (0.61\%)\\
0.0265--0.0414 & 4 / 1,292 (0.31\%) & 52 / 3,190 (1.63\%)\\
$\geq0.0414$ & 1 / 321 (0.31\%) & 201 / 4,161 (4.83\%)\\
\bottomrule
\end{tabular}
\end{minipage}\hfill
\begin{minipage}[t]{0.46\textwidth}
\vspace{0pt}
\centering
\caption{SSR (\%) by tracker-pool size, averaged over all subsets of that size, for CHOREO's rule and for uniform random choice among the eligible trackers. Every SSR divides by 2,000 trials.}
\label{tab:pool-size}
\footnotesize
\begin{tabular}{rrrr}
\toprule
Trackers & Subsets & CHOREO & Random eligible\\
\midrule
1 & 8 & 46.84 & 46.84\\
2 & 28 & 65.55 & 62.53\\
3 & 56 & 75.66 & 69.59\\
4 & 70 & 82.23 & 73.42\\
5 & 56 & 86.95 & 75.64\\
6 & 28 & 90.58 & 76.98\\
7 & 8 & 93.47 & 77.78\\
8 & 1 & 95.80 & 78.26\\
\bottomrule
\end{tabular}
\end{minipage}
\end{table}

\end{document}